\documentclass[journal]{IEEEtran}
\ifCLASSINFOpdf
\else
\fi
\usepackage{palatino,epsfig,latexsym}
\usepackage{booktabs} 
\usepackage{subfigure,cite}
\usepackage{enumitem}
\usepackage{graphicx}
\usepackage{amsthm,amsmath}
\usepackage{amssymb}
\usepackage{color}
\usepackage{url}

\theoremstyle{definition}

\newtheorem{example}{Example}

\newtheorem{property}{Property}

\begin{document}

\title{HV-Net: Hypervolume Approximation based on DeepSets}

\author{Ke~Shang,
	~Weiyu~Chen,
	~Weiduo~Liao,
        ~Hisao~Ishibuchi,~\IEEEmembership{Fellow,~IEEE}
\thanks{This work was supported by National Natural Science Foundation of China (Grant No. 62002152, 61876075), Guangdong Provincial Key Laboratory (Grant No. 2020B121201001), the Program for Guangdong Introducing Innovative and Enterpreneurial Teams (Grant No. 2017ZT07X386), The Stable Support Plan Program of Shenzhen Natural Science Fund (Grant No. 20200925174447003), Shenzhen Science and Technology Program (Grant No. KQTD2016112514355531). \textit{(Corresponding Author: Hisao Ishibuchi.)}}
\thanks{K. Shang, W. Liao, and H. Ishibuchi are with Guangdong Provincial Key Laboratory of Brain-inspired Intelligent Computation, Department of Computer Science and Engineering, Southern University of Science and Technology, Shenzhen 518055, China (e-mail: kshang@foxmail.com; liaowd@mail.sustech.edu.cn; hisao@sustech.edu.cn).}
\thanks{W. Chen is with Department of Computer Science and Engineering, The Hong Kong University of Science and Technology, Hong Kong, China (e-mail: wchenbx@cse.ust.hk).}
\thanks{K. Shang and W. Chen contribute equally to this work.}
}

\maketitle

\begin{abstract}
In this letter, we propose HV-Net, a new method for hypervolume approximation in evolutionary multi-objective optimization. The basic idea of HV-Net is to use DeepSets, a deep neural network with permutation invariant property, to approximate the hypervolume of a non-dominated solution set. The input of HV-Net is a non-dominated solution set in the objective space, and the output is an approximated hypervolume value of this solution set. The performance of HV-Net is evaluated through computational experiments by comparing it with two commonly-used hypervolume approximation methods (i.e., point-based method and line-based method). Our experimental results show that HV-Net outperforms the other two methods in terms of both the approximation error and the runtime, which shows the potential of using deep learning technique for hypervolume approximation.
\end{abstract}

\IEEEpeerreviewmaketitle

\section{Introduction}
In the filed of evolutionary multi-objective optimization (EMO), there are many performance indicators which are used to evaluate the performance of EMO algorithms \cite{li2019quality}. Some representative performance indicators include GD \cite{van1999multiobjective}, IGD \cite{coello2004study}, hypervolume \cite{zitzler2003performance}, R2 \cite{hansen1998evaluating}, etc. Among them, the hypervolume indicator is the most widely investigated one since it has rich theoretical properties and mature applications. For example, it is able to evaluate both the convergence and diversity of a solution set simultaneously \cite{shang2020survey}. It is Pareto compliant \cite{zitzler2007hypervolume}. Furthermore, it can also be used in EMO algorithms for environmental selection. Representative hypervolume-based EMO algorithms include SMS-EMOA \cite{Emmerich2005An,beume2007sms}, FV-MOEA \cite{jiang2014simple}, HypE \cite{bader2011hype}, and R2HCA-EMOA \cite{shang2020new}. 

The main drawback of the hypervolume indicator is that it is computationally more expensive than other performance indicators. Some efficient hypervolume calculation methods have been proposed such as WFG \cite{While2012A}, QHV \cite{Russo2012Quick}, and HBDA \cite{lacour2017box}. However, these methods aim to exactly calculate the hypervolume indicator. They will become inefficient when the number of objectives is large (e.g., $>10$) since the calculation of the hypervolume indicator is \#P-hard \cite{Bringmann2010Approximating}. Therefore, some hypervolume approximation methods have been proposed to overcome this drawback \cite{bader2010faster,Bringmann2010Approximating,ishibuchi2009hypervolume,shangke,deng2019approximating,fieldsend2019efficient,tang2019fast,deng2020combining}.  

Two representative approximation methods are the point-based method and the line-based method. The point-based method is also known as Monte Carlo sampling method \cite{bader2010faster,deng2020combining}. In this method, a large number of points are sampled in the sampling space, and the hypervolume is approximated based on the percentage of the points lying inside the hypervolume region. The line-based method is also known as the R2 indicator method \cite{shangke} or the polar coordinate method \cite{deng2019approximating}. In this method, a set of line segments with different directions are used to approximate the hypervolume. 

In this letter, we propose HV-Net, a new hypervolume approximation method. HV-Net is a deep neural network where the input is a non-dominated solution set and the output is the hypervolume approximation of this solution set. HV-Net has two characteristics. 1) It is permutation invariant to the order of elements in the input set (i.e., the output of HV-Net does not depend on the order of solutions in the input solution set). 2) It can handle an input solution set with an arbitrary number of solutions (i.e., the number of solutions in the input solution set is not fixed). These characteristics make sure that HV-Net is flexible and general for hypervolume approximation. 

The main contribution of this letter is that we develop a new type of hypervolume approximation method based on a deep neural network. Our experimental results are promising and encouraging. This will bring new opportunities for the development of the EMO field.

The rest of this letter is organized as follows. Section II presents the preliminaries of the study. Section III introduces the new hypervolume approximation method, HV-Net. Section IV conducts experimental studies. Section V concludes the letter.

\section{Preliminaries}
\subsection{Hypervolume Indicator}
The hypervolume indicator is a widely used performance indicator in the field of evolutionary multi-objective optimization (EMO). Formally, for a solution set $S$ in the objective space, the hypervolume of $S$ is defined as
\begin{equation}
HV(S,\mathbf{r}) = \mathcal{L}\left(\bigcup_{\mathbf{s}\in S}\left\{\mathbf{s}'|\mathbf{s}\prec \mathbf{s}'\prec \mathbf{r}\right\}\right),
\end{equation}
where $\mathcal{L}(.)$ is the Lebesgue measure of a set, $\mathbf{r}\in\mathbb{R}^m$ is a reference point which is dominated by all solutions in $S$, and $\mathbf{s}\prec \mathbf{s}'$ denotes that $\mathbf{s}$ Pareto dominates $\mathbf{s}'$ (i.e., $s_i\leq s'_i$ for all $i=1,...,m$ and $s_j<s'_j$ for at least one $j=1,...,m$ in the minimization case, where $m$ is the number of objectives).

\begin{figure}[!htbp]
\centering                                           
\includegraphics[scale=0.33]{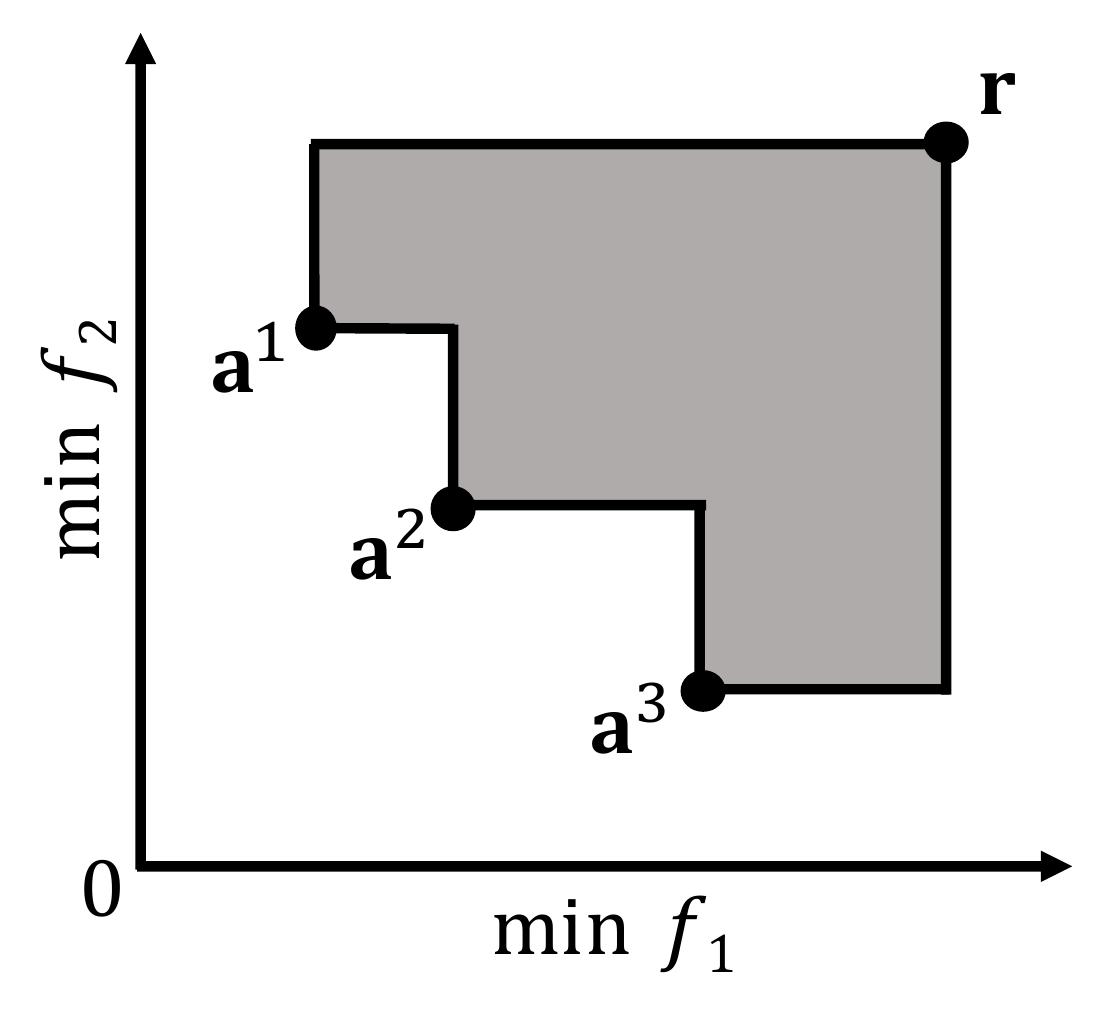}            
\caption{An illustration of the hypervolume indicator. The shaded area is the hypervolume of the solution set $S= \{\mathbf{a}^1,\mathbf{a}^2,\mathbf{a}^3\}$ based on reference point $\mathbf{r}$, i.e., $HV(S,\mathbf{r})$. } 
\label{hv}                                                        
\end{figure}

Fig. \ref{hv} gives an illustration of the hypervolume of a solution set $S=\{\mathbf{a}^1,\mathbf{a}^2,\mathbf{a}^3\}$ in a two-dimensional objective space where each objective is to be minimized. 

\subsection{Hypervolume Approximation Methods}
The main drawback of the hypervolume indicator is that it is computationally expensive in high-dimensional spaces \cite{Bringmann2010Approximating}, which limits its applicability to many-objective optimization. In order to overcome this drawback, some hypervolume approximation methods have been proposed \cite{bader2010faster,Bringmann2010Approximating,ishibuchi2009hypervolume,shangke,deng2019approximating,fieldsend2019efficient,tang2019fast,deng2020combining}. Two representative methods are the point-based method and the line-based method. These two methods are briefly explained bellow.

\subsubsection{Point-based method}
The point-based method is also known as Monte Carlo sampling method \cite{bader2010faster,deng2020combining}. In this method, a large number of points are sampled in the sampling space and the hypervolume is approximated based on the percentage of the points lying inside the hypervolume region. 

\begin{figure}[!htbp]
\centering                                           
\subfigure[Point-based method]{ 
\includegraphics[scale=0.33]{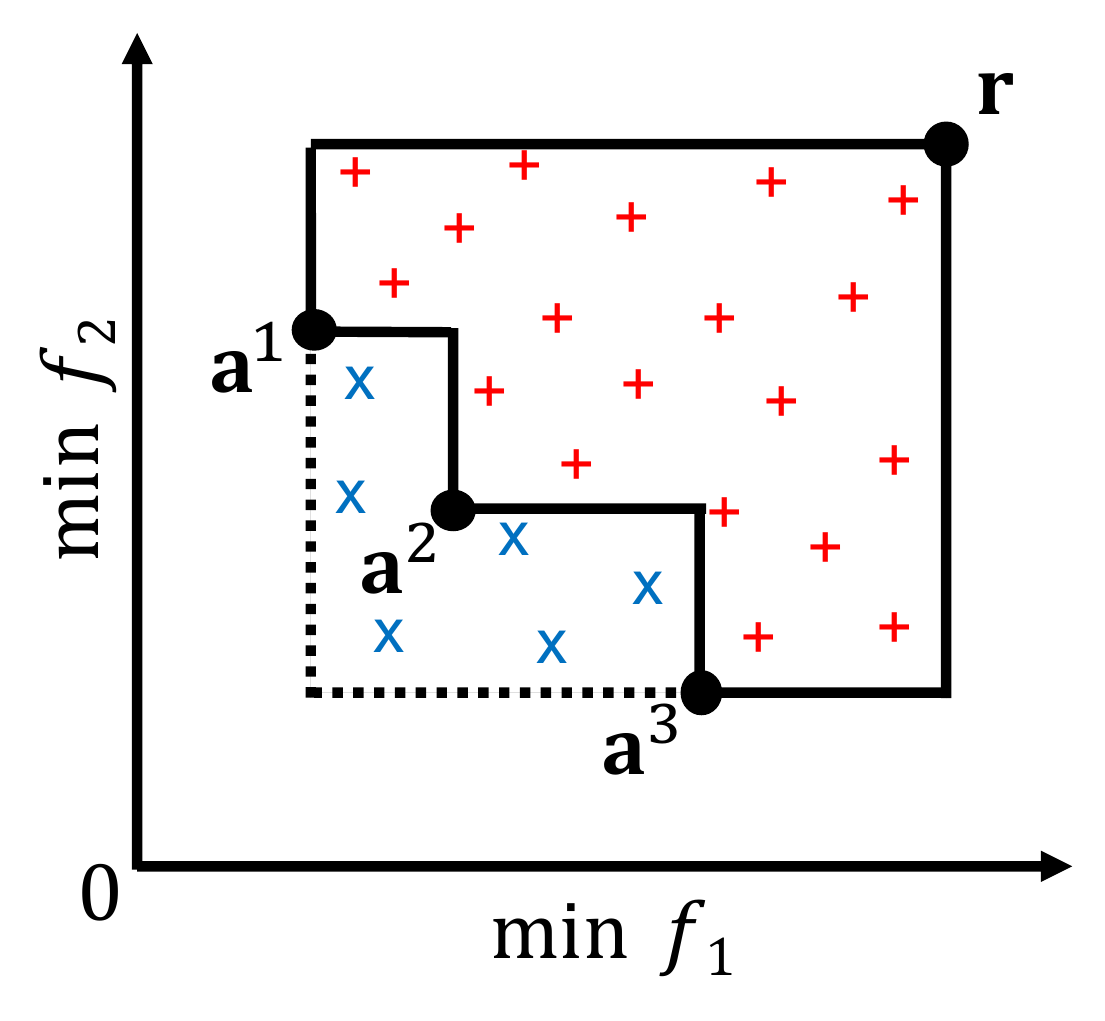}   }            
\subfigure[Line-based method]{ 
\includegraphics[scale=0.33]{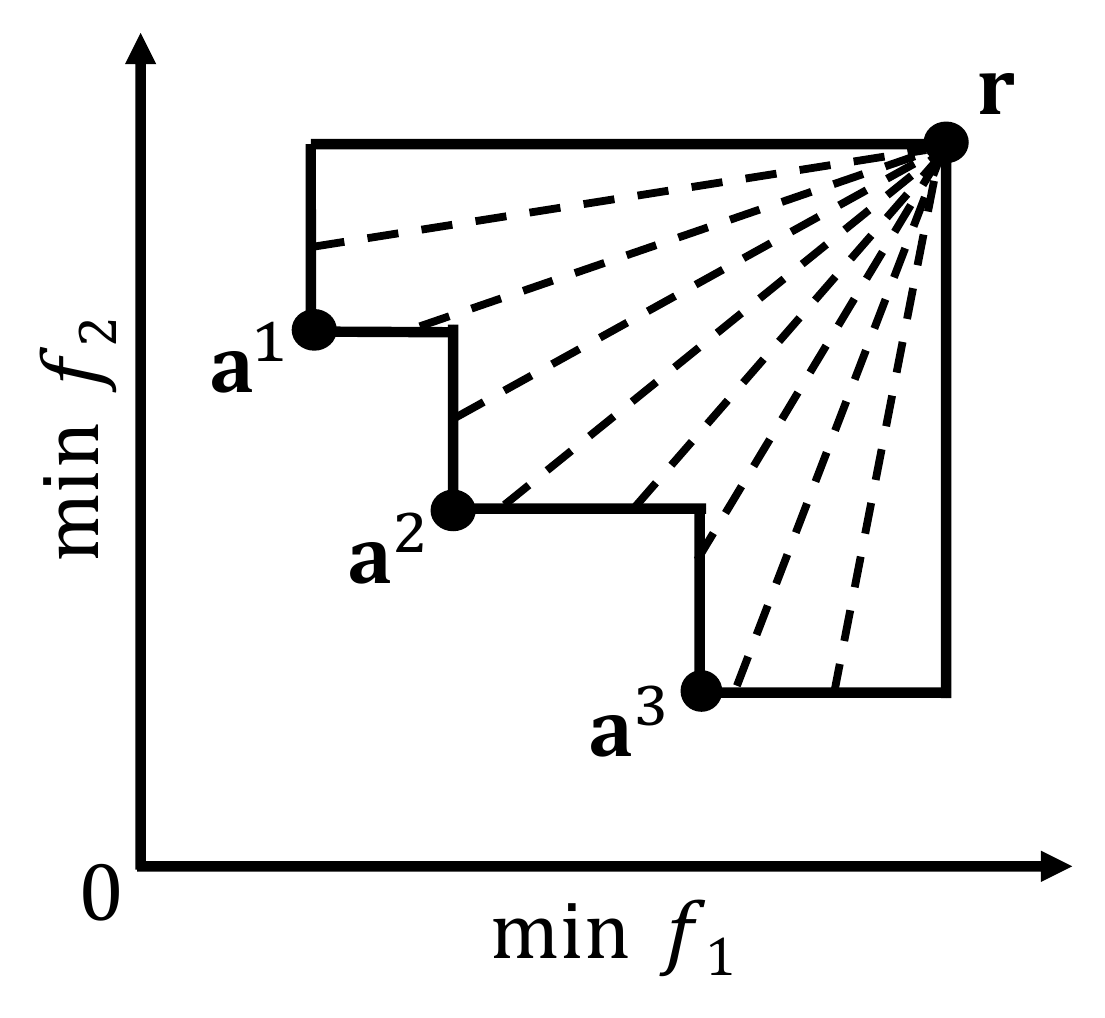} }
\caption{Illustration of the hypervolume approximation methods.} 
\label{hvapprox}                                                        
\end{figure}

Fig. \ref{hvapprox} (a) illustrates this method. The sampling space is determined by the reference point and the ideal point of the solution set $S$. The volume of the sampling space can be easily calculated since it is a cuboid. We assume the volume of the sampling space is $V$. Then $k$ points are uniformly sampled in the sampling space. Suppose that there are $k'$ points dominated by $S$ (i.e., these points lie in the hypervolume region of $S$), then the hypervolume of $S$ is approximated as
\begin{equation}
{HV}(S,\mathbf{r})\approx \frac{k'}{k}V.
\end{equation}

\subsubsection{Line-based method}
The line-based method is also known as the R2 indicator method \cite{shangke} or the polar coordinate method \cite{deng2019approximating}. In this method, a set of line segments with different directions are used to approximate the hypervolume. 

Fig. \ref{hvapprox} (b) illustrates this method. A set of line segments with different directions are drawn from the reference point to the boundary of the hypervolume region. Suppose that we have $n$ line segments and the length of each line segment in the hypervolume region is $l_i,i=1,...,n$, then the hypervolume of $S$ is approximated as 
\begin{equation}
\label{lineapprox1}
{HV}(S,\mathbf{r})\approx \frac{\pi^{m/2}}{mn2^{m-1}\Gamma(m/2)}\sum_{i=1}^n l_i^m,
\end{equation}
where $\Gamma(x) = \int_{0}^{\infty}z^{x-1}e^{-z}dz$ is the Gamma function.

The directions of the line segments can be defined using a direction vector set $\Lambda = \{\boldsymbol{\lambda}^1,...,\boldsymbol{\lambda}^n\}$ where each direction vector satisfies $\left \| {\boldsymbol{\lambda}^i} \right \|_2 = 1$, $\lambda^i_j\geq 0$, $i=1,...,n$, $j=1,...,m$. The length of each line segment in the hypervolume region can be calculated as
\begin{equation}
\label{lineapprox2}
l_i = \max_{\mathbf{s}\in S}\min_{j\in\{1,...,m\}}\left\{\frac{|r_j-s_j|}{\lambda^i_j}\right\}, i=1,...,n.
\end{equation}

For more information about the derivation of \eqref{lineapprox1} and \eqref{lineapprox2}, please refer to \cite{shangke,deng2019approximating}.

\section{HV-Net}
\begin{figure}[!htbp]
\centering                                           
\includegraphics[scale=0.35]{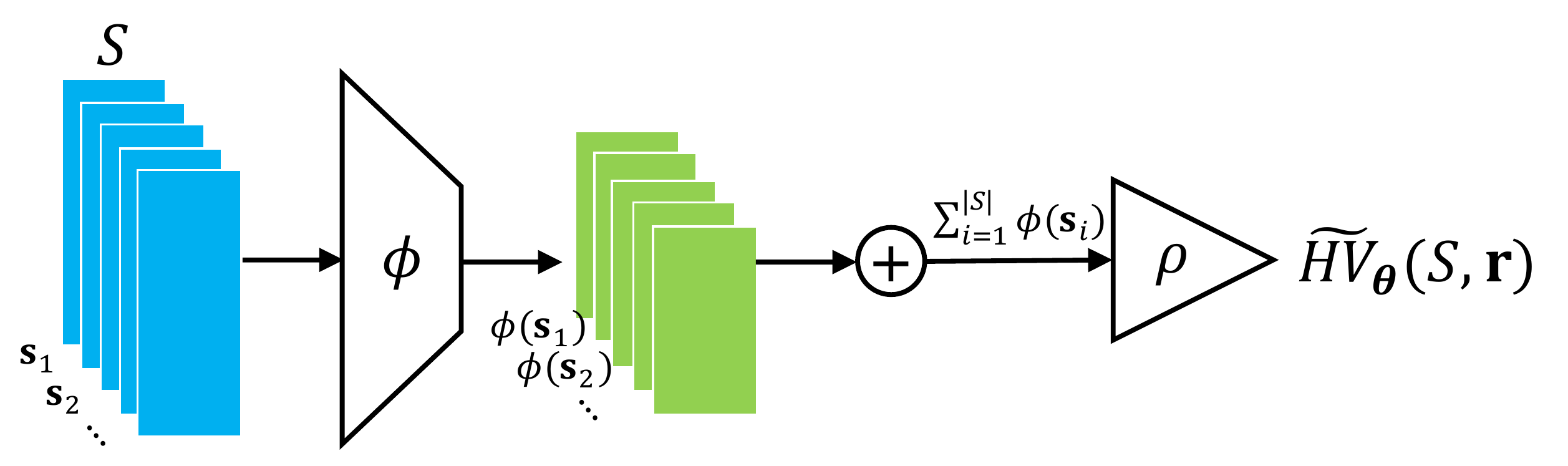}            
\caption{The architecture of HV-Net.} 
\label{hvnet}                                                        
\end{figure}

In this section, we propose a new type of hypervolume approximation method, HV-Net. HV-Net is based on DeepSets \cite{zaheer2017deep}, which is a fundamental architecture to deal with sets as inputs. The architecture of HV-Net is shown in Fig. \ref{hvnet}. The input of HV-Net is a non-dominated solution set $S=\{\mathbf{s}_1,\mathbf{s}_2,...,\mathbf{s}_N\}$. First, each of $N$ solutions $\mathbf{s}_i\in S$ is presented to the network $\phi$ and transformed to $\phi(\mathbf{s}_i)$. Then, these $N$ transformed vectors are added up as one vector $\sum_{i=1}^N \phi(\mathbf{s}_i)$. This vector is further presented to another network $\rho$, and the hypervolume of $S$ is approximated as $\widetilde{HV}_{\boldsymbol{\theta}}(S,\mathbf{r})= \rho\left (\sum_{i=1}^N \phi(\mathbf{s}_i)\right )$ where $\boldsymbol{\theta}$ is the parameter vector of HV-Net.

The most important property of HV-Net is that it is permutation invariant to the order of elements in the set (i.e., the sequence of solutions in $S$ in this paper). That is, the output of HV-Net is the same for any two sets as long as they have the same elements. Another property of HV-Net is that it can handle any solution set with an arbitrary number of solutions, which is important since we usually need to approximate the hypervolume of many solution sets with different number of solutions, and we want to use a single neural network to do this. These two properties make sure that HV-Net is suitable and flexible for hypervolume approximation. 

In the next subsections, we discuss in detail how to train and use HV-Net for hypervolume approximation. 

\subsection{How to Train HV-Net}
\label{trainhvnet}
In HV-Net, we implicitly assume the minimization case where the reference point for hypervolume calculation is set to $\mathbf{r}=(1,...,1)$, and all solutions in $S$ are located in $[0,1]^m$. For the training of HV-Net, we prepare the training data as follows. First, we prepare $L$ non-dominated solution sets $\{S_1,S_2,...,S_L\}$ where each solution set is located in $[0,1]^m$. Then, we calculate the hypervolume of each solution set based on the reference point $\mathbf{r}=(1,...,1)$. That is, we obtain the target output $HV (S_i, \mathbf{r})$ for each solution set $S_i$ $(i = 1, 2, ..., L)$ for the learning of HV-Net. For hypervolume calculation, we use the WFG algorithm \cite{While2012A}.


Based on the training data (i.e., $L$ input-ouput pairs $(S_i, HV(S_i, \mathbf{r}))$, $i = 1, 2, ..., L$), we define the loss function of HV-Net as follows:
\begin{equation}
\label{lossfunc}
\mathcal{L}(\boldsymbol{\theta}) = \frac{1}{L}\sum_{i=1}^L \left(\log\widetilde{HV}_{\boldsymbol{\theta}}(S_i,\mathbf{r})-\log HV(S_i,\mathbf{r})\right)^2.
\end{equation}

We will discuss in detail why we use \eqref{lossfunc} as the loss function for training in Section \ref{2lossfuncs}. Based on \eqref{lossfunc}, our target is to find the optimal parameter vector $\mathbf{\boldsymbol{\theta}}^*=\arg\min \mathcal{L}(\boldsymbol{\theta})$ for HV-Net. More details about the training of HV-Net are provided in Section \ref{netspec}.

\subsection{How to Use HV-Net} 
Suppose that we have a well-trained HV-Net. The question is how to use it for hypervolume approximation when the solution set and the reference point are both arbitrarily given. The basic idea is to first transform the solution set and the reference point so that the reference point is $\mathbf{r}=(1,...,1)$ and the solution set is located in $[0,1]^m$. Then we use HV-Net to approximate the hypervolume of the transformed solution set. Lastly, we calculate the hypervolume approximation of the original solution set based on the output of HV-Net.

We realize the above described procedure based on the following properties of the hypervolume indicator. To the best of our knowledge, this is the first study to explicitly show the following properties for the hypervolume indicator in the literature. 
\begin{property}
For any positive vector $\boldsymbol{\alpha}\in \mathbb{R}^m_{>0}$, $HV(S,\mathbf{r}) = \frac{1}{\prod_{i=1}^m \alpha_i }HV(\boldsymbol{\alpha}\odot S, \boldsymbol{\alpha}\odot\mathbf{r})$, where $\odot$ denotes the element-wise multiplication\footnote{For two vectors $\mathbf{a}=(a_1,...,a_m)$ and $\mathbf{b}=(b_1,...,b_m)$, $\mathbf{a}\odot\mathbf{b} = (a_1b_1,...,a_mb_m)$. For a set $B$, $\mathbf{a}\odot B$ means $\mathbf{a}\odot\mathbf{b}$ for all $\mathbf{b}\in B$.}.
\end{property}
\begin{property}
For any real vector $\boldsymbol{\beta}\in \mathbb{R}^m$, $HV(S,\mathbf{r}) = HV(S+\boldsymbol{\beta},\mathbf{r}+\boldsymbol{\beta})$.
\end{property}
\begin{property}
$HV(S,\mathbf{r}) = HV(-S, -\mathbf{r})$ where $HV(-S, -\mathbf{r})$ is calculated for maximization problems whereas $HV(S, \mathbf{r})$ is calculated for minimization problems.
\end{property}

The above three properties can be easily derived based on the properties of the Lebesgue measure \cite{enwiki:1061315833}. Using the above three properties, we can transform any solution set and any reference point so that they meet the requirement of HV-Net. Thus we can calculate the hypervolume approximation of the original solution set. Next, we give a toy example to illustrate how to use HV-Net for hypervolume approximation.
\begin{example}
Suppose that we have a solution set $S=\{(10,1),(7,3),(4,7)\}$, and the reference point $\mathbf{r} = (11,9)$. First, we transform the solution set and the reference point to $S'=\{(10-4,1-1),(7-4,3-1),(4-4,7-1)\} $ and $\mathbf{r}' = (11-4,9-1)$ so that the minimum value of each objective becomes zero. Based on \textbf{Property 2}, we know that $HV(S,\mathbf{r})=HV(S',\mathbf{r}')$. Then we transform the solution set and the reference point to $S''=\{(6/7,0/8),(3/7,2/8),(0/7,6/8)\} $ and $\mathbf{r}'' = (1,1)$ so that the reference point becomes $(1, 1)$. Based on \textbf{Property 1}, we know that $HV(S',\mathbf{r}')=7\times 8\times HV(S'',\mathbf{r}'')$. Now we can use HV-Net to approximate the hypervolume of $S''$ as $\widetilde{HV}_{\boldsymbol{\theta}}(S'',\mathbf{r}'' )$. The hypervolume of $S$ is approximated as $HV(S,\mathbf{r}) =56\times HV(S'',\mathbf{r}'' ) \approx 56\times \widetilde{HV}_{\boldsymbol{\theta}}(S'',\mathbf{r}'' )$. 
\end{example}

\textbf{Property 3} is used to deal with the maximization case for hypervolume calculation. That is, we can transform maximization to minimization based on \textbf{Property 3}.

\section{Experiments}
In this section, we evaluate the performance of HV-Net by comparing it with the other two hypervolume approximation methods (i.e., the point-based and line-based methods).

\subsection{Experimental Settings}
\label{netspec}
\subsubsection{HV-Net specifications}
\begin{figure}[!htbp]
\centering                                           
\subfigure[Network $\phi$]{ 
\includegraphics[scale=0.35]{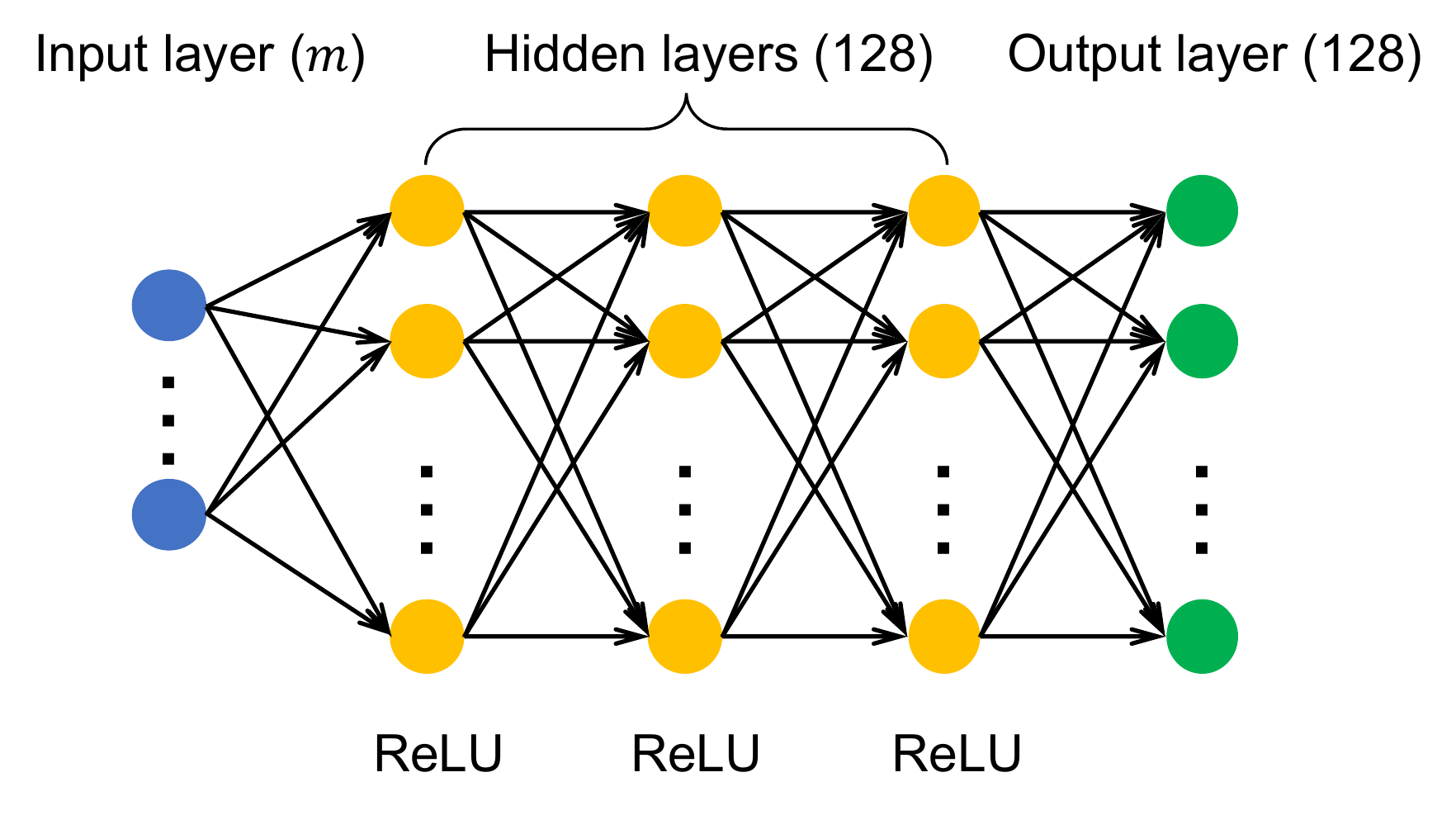}   }            
\subfigure[Network $\rho$]{ 
\includegraphics[scale=0.35]{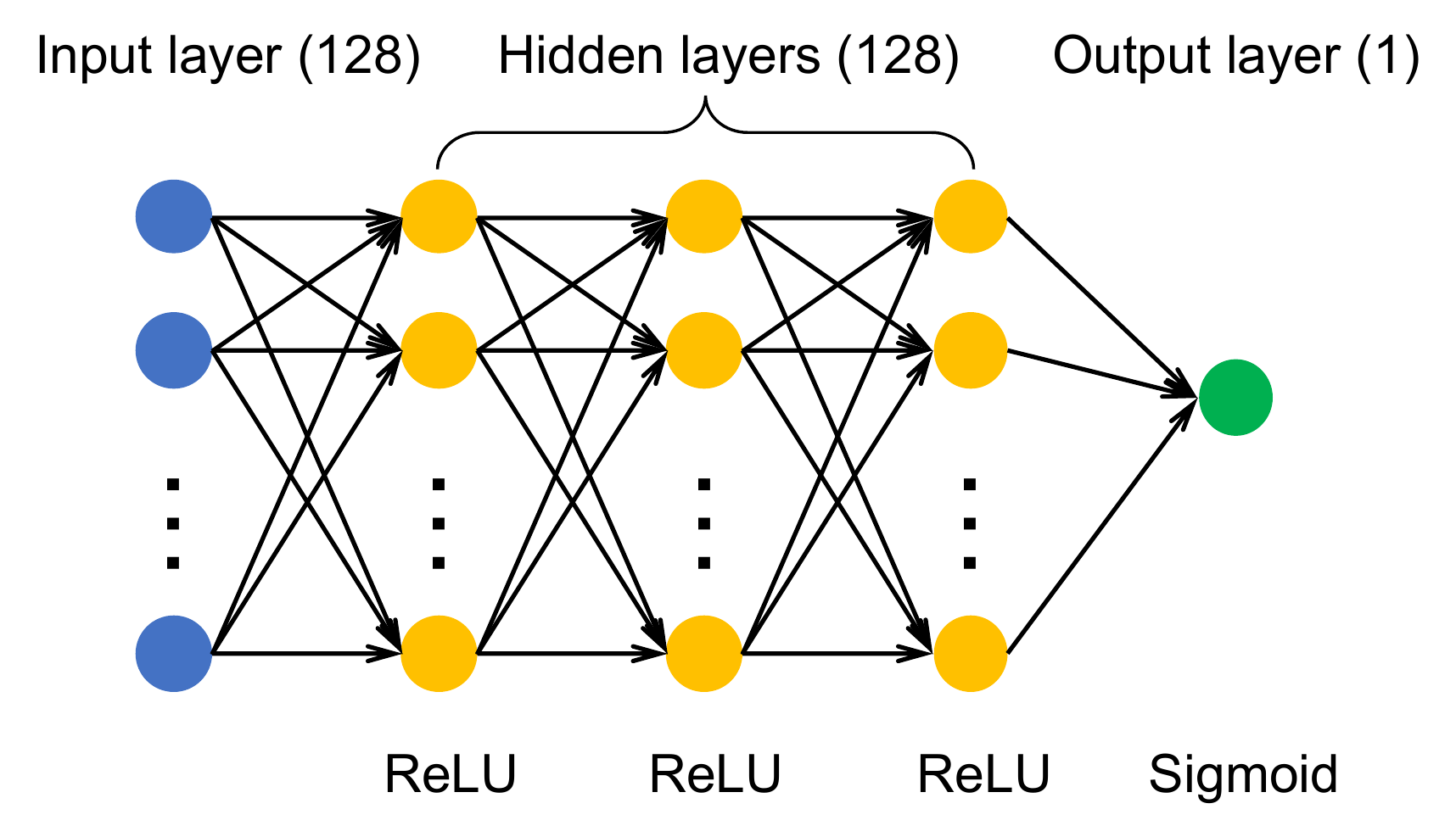} }
\caption{Networks $\phi$ and $\rho$ in HV-Net. The number in the parentheses indicates the number of neurons in each layer. The activation function used in each layer is shown under each layer.} 
\label{networks}                                                        
\end{figure}

In HV-Net in Fig. \ref{hvnet}, two networks $\phi$ and $\rho$ need to be specified. In our experiments, $\phi$ and $\rho$ are specified as two feedforward neural networks. Both of them have three hidden layers where each hidden layer has 128 neurons. The input dimension of $\phi$ is set to $m$ (i.e., the number of objectives). The output dimension of $\phi$ is set to 128. The input dimension of $\rho$ is the same as the output dimension of $\phi$ (i.e., 128). The output dimension of $\rho$ is set to 1. Fig. \ref{networks} shows the structures of $\phi$ and $\rho$.

\subsubsection{Training and testing data generation}
\label{traindatagen}
We consider 3, 5, 8, and 10-objective cases (i.e., $m=3,5,8,10$). To generate training solution sets $\{S_1,S_2,...,S_L\}$ for each case, we set $L=1,000,000$. Each solution set is generated using the following procedure: 
\begin{itemize}
\item Step 1: Randomly sample an integer $num\in [1,100]$ where $num$ denotes the number of solutions in the solution set.
\item Step 2: Randomly sample 1000 solutions in $[0,1]^m$ as candidate solutions.
\item Step 3: Apply non-dominated sorting to these 1000 solutions and obtain different fronts $\{F_1,F_2,...\}$ where $F_1$ is the first front (i.e., the set of non-dominated solutions in the 1000 solutions) and $F_2$ is the set of non-dominated solutions after all solutions in $F_1$ are removed.
\item Step 4: Start from $F_1$ until the front $F_l$ with no less than $num$ solutions is found. If there is no front satisfies this condition, go back to Step 2.
\item Step 5: Randomly select $num$ solutions from the front $F_l$ to construct one solution set.
\end{itemize} 

This procedure is used in order to select a wide variety of non-dominated solution sets. On average, about 10,000 solution sets with the same size are generated (1,000,000 solution sets in total for 100 different sizes). Some statistical information of the training solution sets is provided in Section I of the supplementary material. Readers can refer there to gain more insights on the training solution sets.

We also use the above procedure to generate testing solution sets. The number of testing solution sets is 10,000 for each objective case. These 10,000 solution sets form one group. We generate a total number of 20 different groups for each objective case.  




\subsubsection{Parameter settings}
For the training of HV-Net, we use Adam \cite{kingma2015adam}, an effective gradient-based optimization method with an adaptive learning rate. The learning rate is set to $10^{-4}$. For all the other parameters in Adam, we use their default settings in PyTorch \cite{paszke2019pytorch}. The batch size during training is set to 100. The number of epochs for training is set to 100.

As the number of sampling points in the point-based method, we examine 20 different specifications: 100, 200, ..., 2000. As the number of lines in the line-based method, we examine 20 different specifications: 10, 20, ..., 200. To generate the direction vector set $\Lambda$ used in the line-based method, we use the unit normal vector (UNV) method \cite{deng2019approximating} since it has a good approximation quality among different direction vector set generation methods \cite{nan2020good}.

\subsubsection{Performance metrics}
To compare the performance of different hypervolume approximation methods, we use the approximation error which is defined as follows:
\begin{equation}
\label{approerror}
\varepsilon = \left |\frac{\widetilde{HV}(S,\mathbf{r})-{HV}(S,\mathbf{r})}{HV(S,\mathbf{r})}\right |,
\end{equation}
where $\widetilde{HV}(S,\mathbf{r})$ denotes the hypervolume approximation. The smaller the approximation error, the better the approximation quality of a method. 

We also record the runtime of the three methods to compare their efficiency. Here the runtime of HV-Net means the evaluation time on the testing solution sets, not its training time.

\subsubsection{Platforms}
All the methods are coded in Python and tested on a server with Intel(R) Xeon(R) Gold 6130 CPU @ 2.10GHz, GeForce RTX 2080 GPU, and Ubuntu 18.04.6 LTS. HV-Net is implemented based on PyTorch version 1.9.0.

\subsection{The Training of HV-Net}
\begin{figure}[!htbp]
\centering                                           
\subfigure[3-objective]{ 
\includegraphics[scale=0.27]{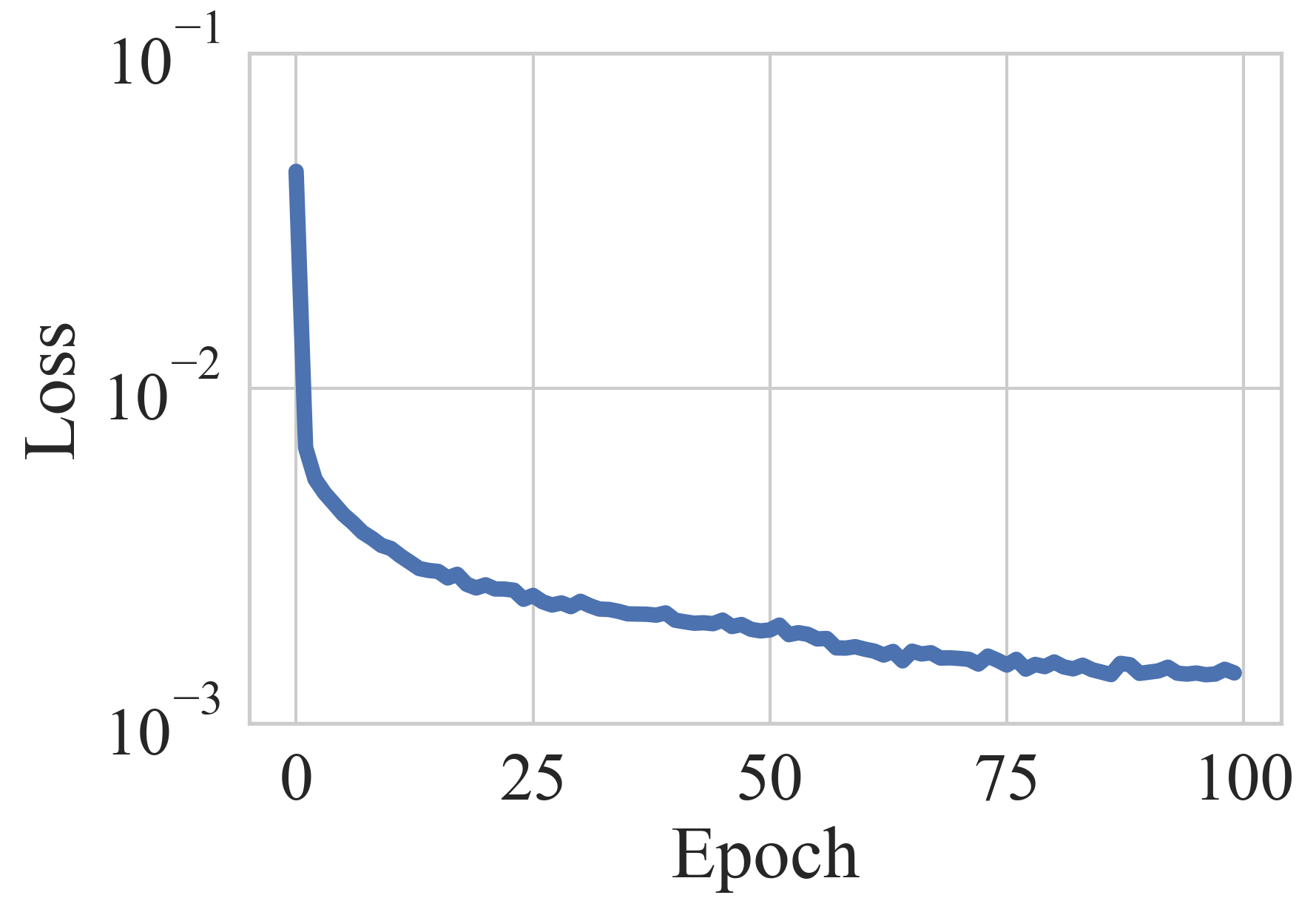}   }            
\subfigure[5-objective]{ 
\includegraphics[scale=0.27]{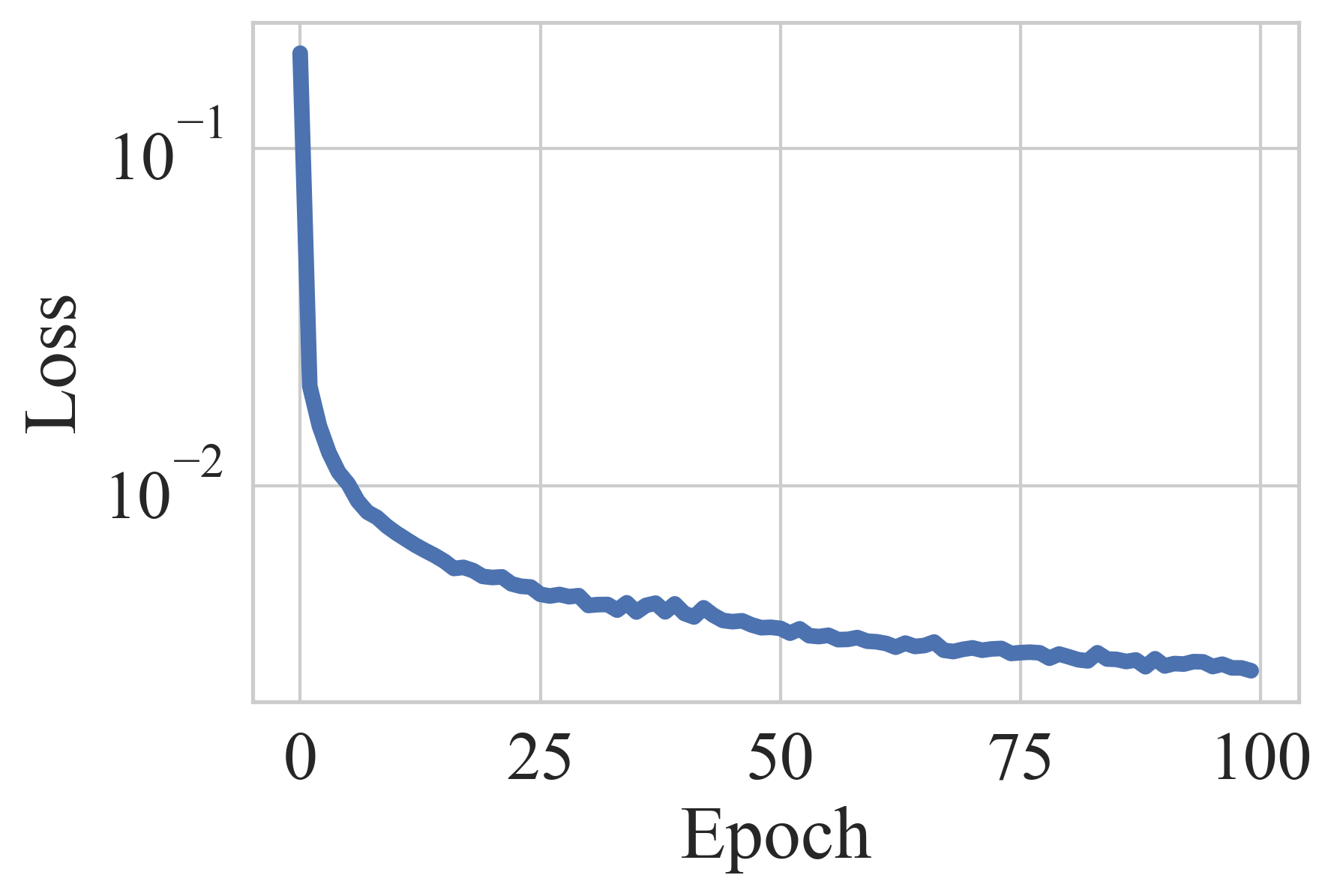} }
\subfigure[8-objective]{ 
\includegraphics[scale=0.27]{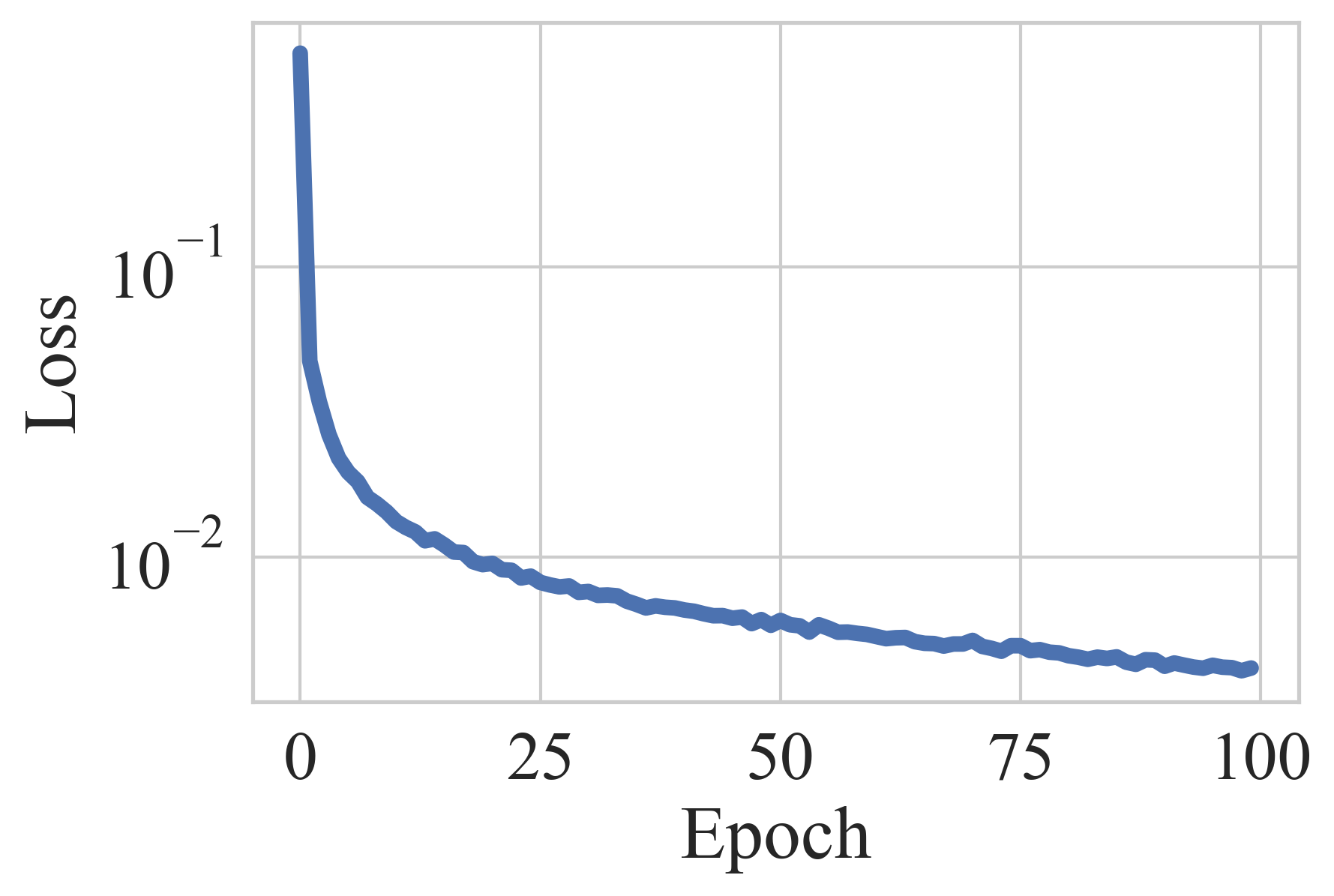} }
\subfigure[10-objective]{
\includegraphics[scale=0.27]{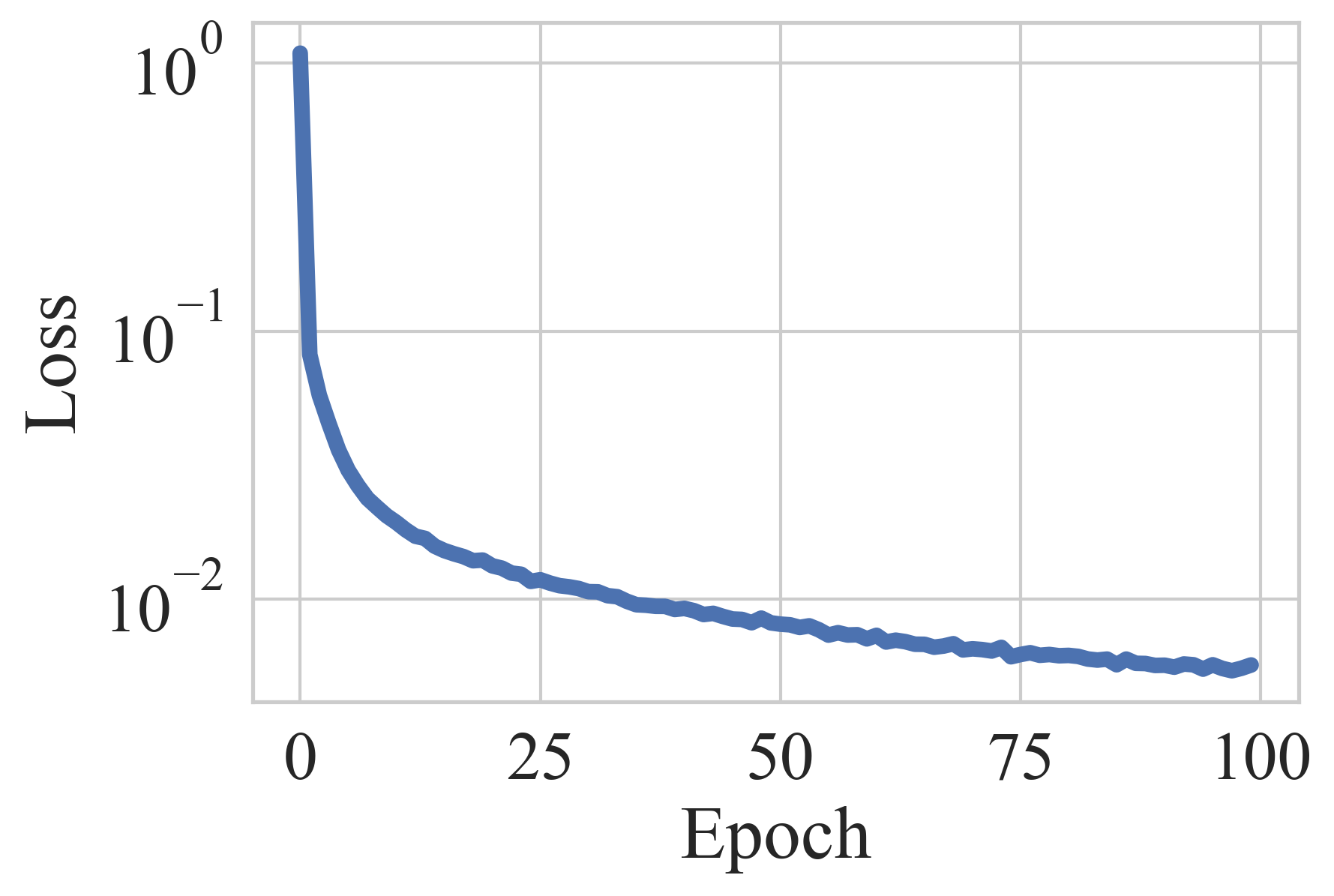} }
\caption{The training curve of HV-Net in each objective case.} 
\label{train}                                                        
\end{figure}

Fig. \ref{train} shows the training curve of HV-Net in each objective case. We can see that the loss becomes very small at the end of the training process in each objective case, which shows the success of the training of HV-Net.

Table \ref{traintime} shows the computation time used for training HV-Net. All the training can be completed within 40 GPU hours.

\begin{table}[!htb]
\centering
\caption{The time (GPU hours) used for training HV-Net in each case.}
\renewcommand\arraystretch{1.5}
\begin{tabular*}{\linewidth}{p{1.9cm}p{1.9cm}p{1.9cm}p{1.9cm}p{1cm}}
\toprule
3-objective & 5-objective & 8-objective & 10-objective  \\\hline
39.077 &39.017 &35.820 &37.930 \\\bottomrule
\end{tabular*}
\label{traintime}
\end{table}

\subsection{Performance Comparison}
We apply the three hypervolume approximation methods on the testing solution sets. For fair comparison, all the methods are tested on CPU. That is, we disable GPU when using HV-Net for evaluation.

\begin{figure}[!htbp]
\centering                                           
\includegraphics[scale=0.6]{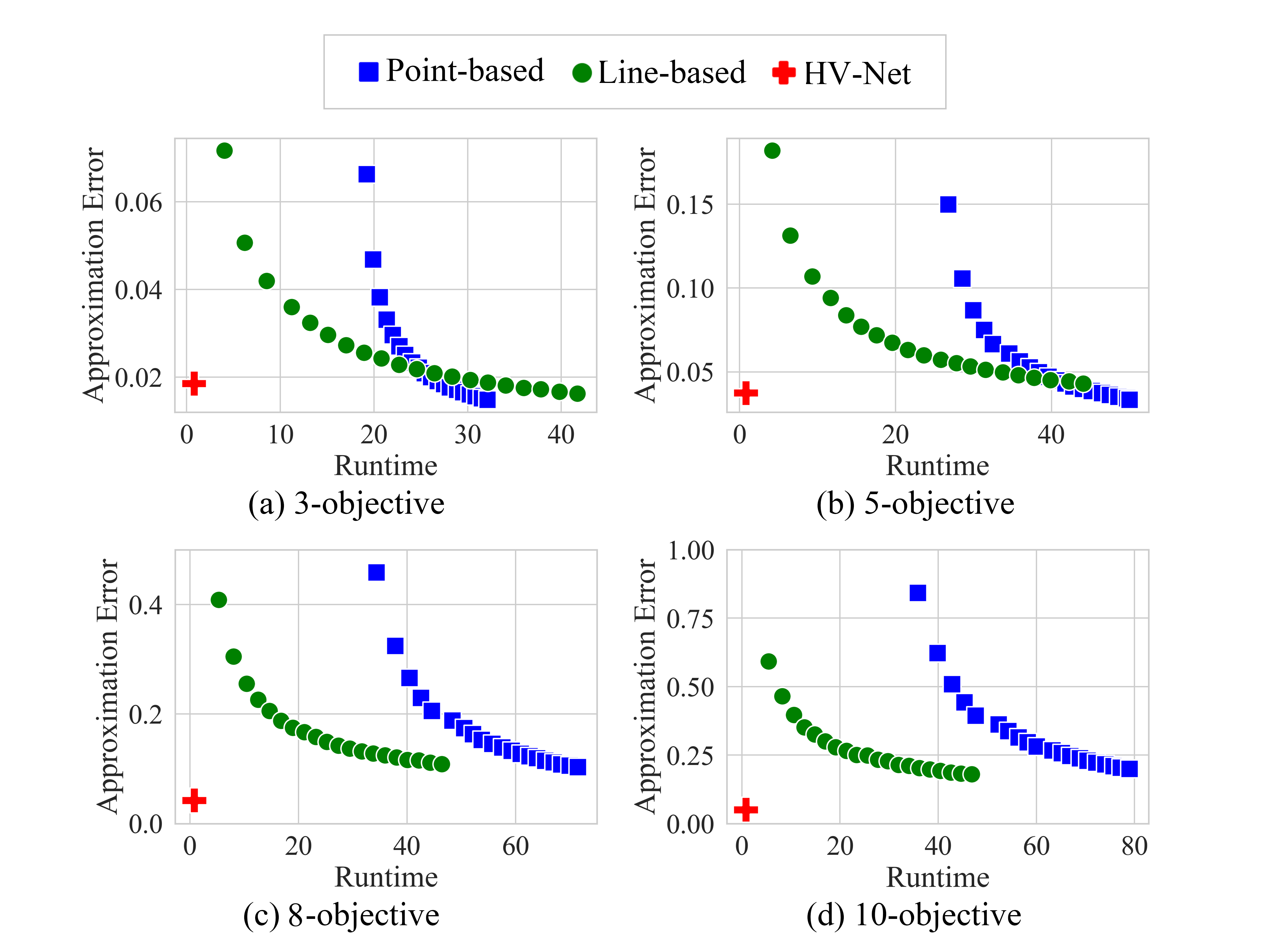} 
\caption{Comparison results of the three hypervolume approximation methods on the testing solution sets in each case. The runtime means the total time for evaluating 10,000 testing solution sets. The approximation error means the average error for 10,000 testing solution sets. All the results are the average over 20 groups of testing solution sets.} 
\label{results}                                                        
\end{figure}

Fig. \ref{results} shows the comparison results for each case. We can see that HV-Net clearly dominates the other two methods in most cases in terms of both the approximation error and the runtime, which shows the advantage of using HV-Net for hypervolume approximation. 

However, we can also see that only a single point is obtained for HV-Net in each case. This is because the structure of HV-Net is fixed. We cannot obtain a set of tradeoff points in each case. For example, we cannot choose another HV-Net for further improving the approximation quality by using more computation time (whereas the computation time can be adjusted in the other methods through the number of points and lines). This is one disadvantage of HV-Net compared with the other two methods.
 
\subsection{Further Investigations}
\label{2lossfuncs}
In Section \ref{trainhvnet}, we defined the loss function $\mathcal{L}$ in \eqref{lossfunc} for HV-Net. Now, we further investigate the following two loss functions:
\begin{equation}
\label{lossfunc1}
\begin{aligned}
&\mathcal{L}_1(\boldsymbol{\theta}) = \frac{1}{L}\sum_{i=1}^L \left(\widetilde{HV}_{\boldsymbol{\theta}}(S_i,\mathbf{r})- HV(S_i,\mathbf{r})\right)^2,\\
&\mathcal{L}_2(\boldsymbol{\theta}) = \frac{1}{L}\sum_{i=1}^L \left |\frac{\widetilde{HV}_{\boldsymbol{\theta}}(S_i,\mathbf{r})-HV(S_i,\mathbf{r})}{HV(S_i,\mathbf{r})}\right |,
\end{aligned}
\end{equation}
where $\mathcal{L}_1$ is the standard mean squared error (MSE) loss function, and $\mathcal{L}_2$ is the mean absolute percentage error (MAPE) loss function \cite{de2016mean}. 

We investigate $\mathcal{L}_1$ since it is more commonly used as the loss function, and $\mathcal{L}_2$ since it is directly related to the approximation error in \eqref{approerror}. We follow the experimental settings described in Section \ref{netspec}. The HV-Net trained by $\mathcal{L}_1$ and $\mathcal{L}_2$ are denoted as HV-Net1 and HV-Net2, respectively. 

\begin{figure}[!htbp]
\centering                                           
\includegraphics[scale=0.6]{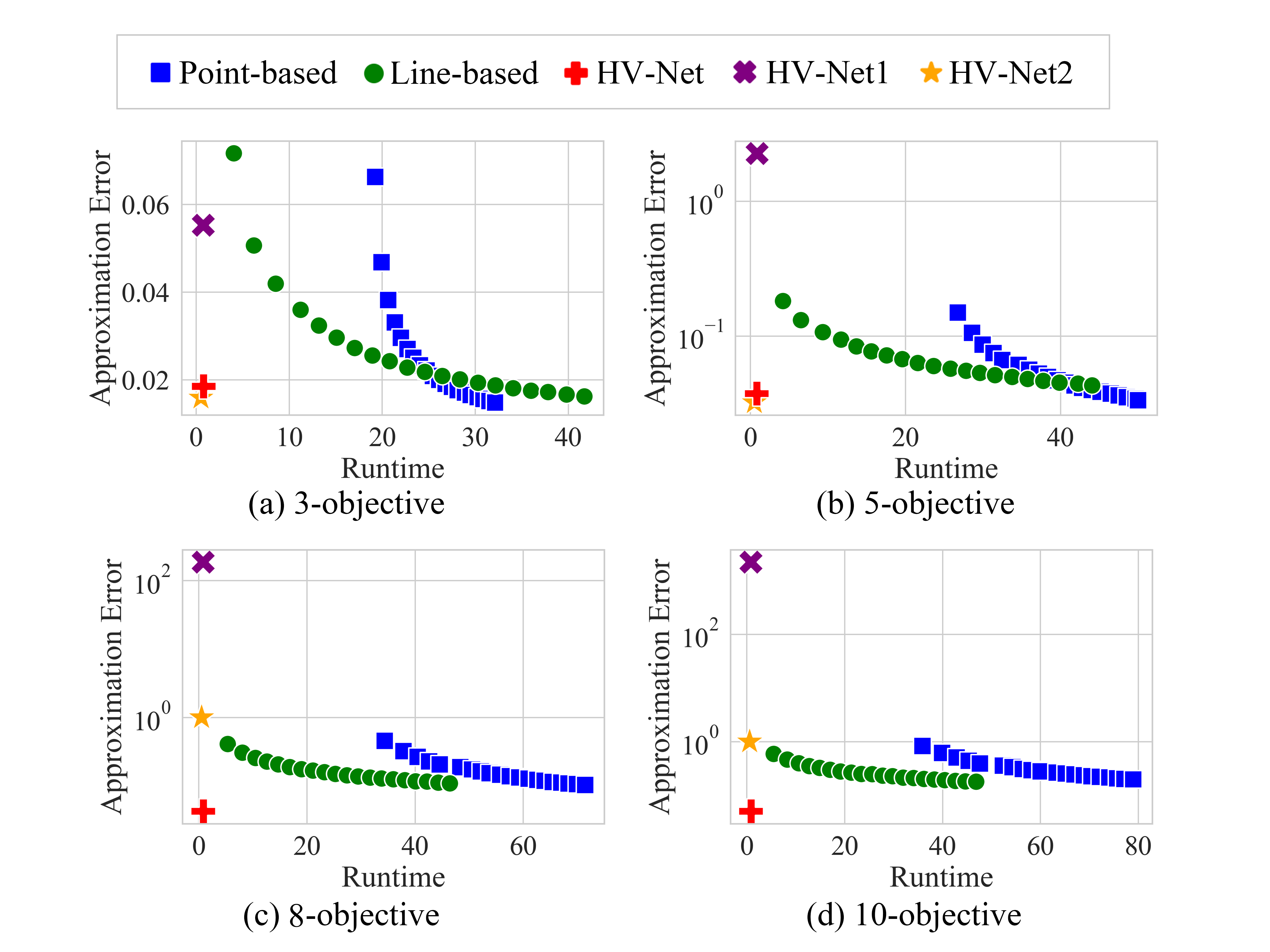}
\caption{Comparison results of HV-Net, HV-Net1, and HV-Net2 on the testing solution sets in each case.} 
\label{2lossfuns}                                                        
\end{figure}

\begin{table}[!htb]
\centering
\caption{The average approximation error achieved by the three HV-Nets on the testing solution sets in each case.}
\renewcommand\arraystretch{1.5}
\begin{tabular*}{\linewidth}{p{1.1cm}p{2.2cm}p{2.2cm}p{2.2cm}p{1cm}}
\toprule
$m$ & HV-Net & HV-Net1 & HV-Net2  \\\hline
3 &0.018541 &0.055279 &0.015878 \\\hline
5 &0.037459 &2.270159 &0.032067  \\\hline
8 &0.042566 &187.189718 &1.000000  \\\hline
10 &0.050867 &2212.552783&1.000000 \\\bottomrule
\end{tabular*}
\label{table1}
\end{table}

\begin{table}[!htb]
\centering
\caption{The average approximation error achieved by the three HV-Nets on the training solution sets in each case.}
\renewcommand\arraystretch{1.5}
\begin{tabular*}{\linewidth}{p{1.1cm}p{2.2cm}p{2.2cm}p{2.2cm}p{1cm}}
\toprule
$m$ & HV-Net & HV-Net1 & HV-Net2  \\\hline
3 &0.018489 &0.056103 &0.015699 \\\hline
5 &0.030241 &4.777839 &0.030650  \\\hline
8 &0.036301 &1094.425093 &1.000000  \\\hline
10 &0.042178 &6848.726503&1.000000 \\\bottomrule
\end{tabular*}
\label{table2}
\end{table}

\begin{figure}[!htbp]
\centering                                           
\subfigure[3-objective]{ 
\includegraphics[scale=0.27]{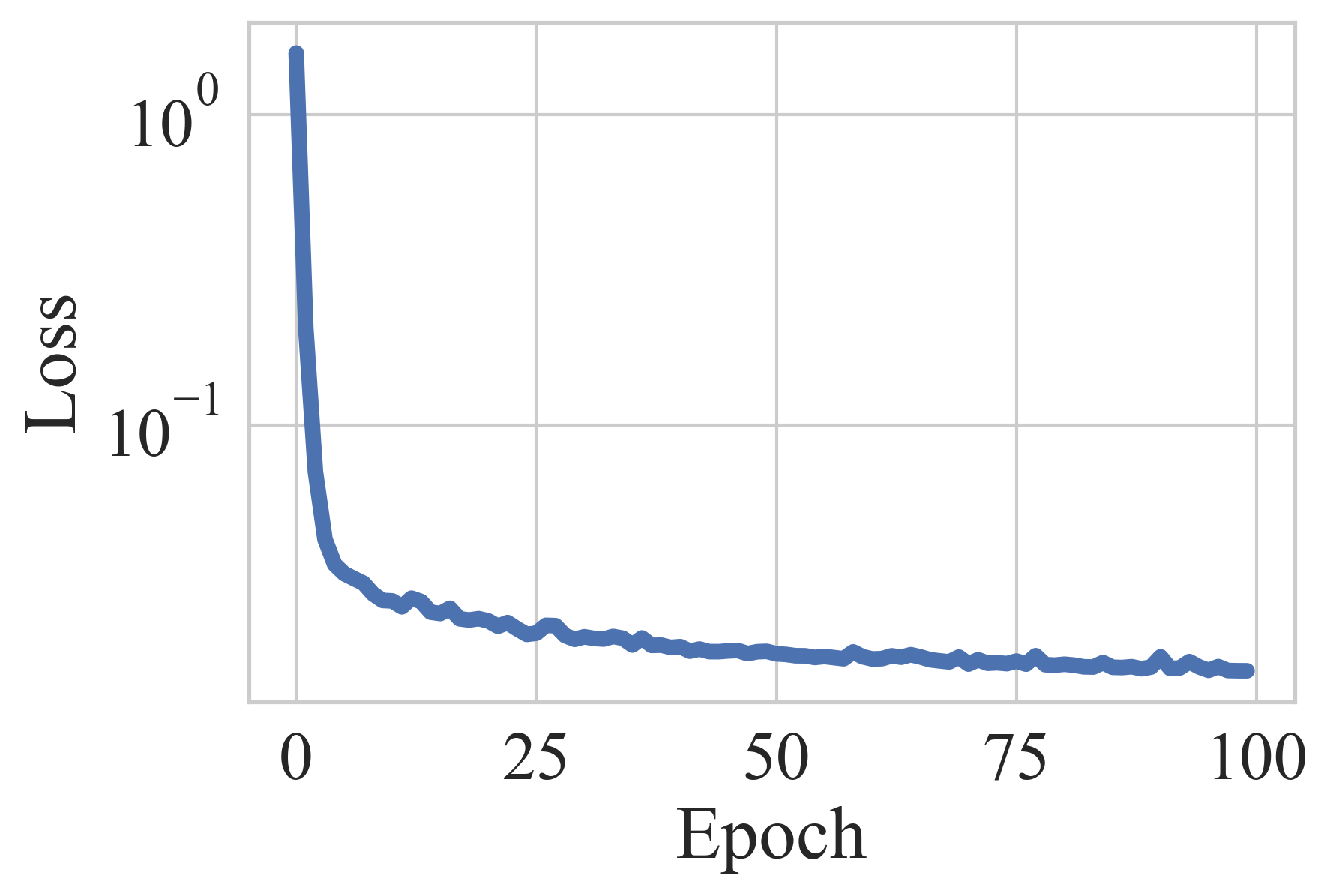}   }            
\subfigure[5-objective]{ 
\includegraphics[scale=0.27]{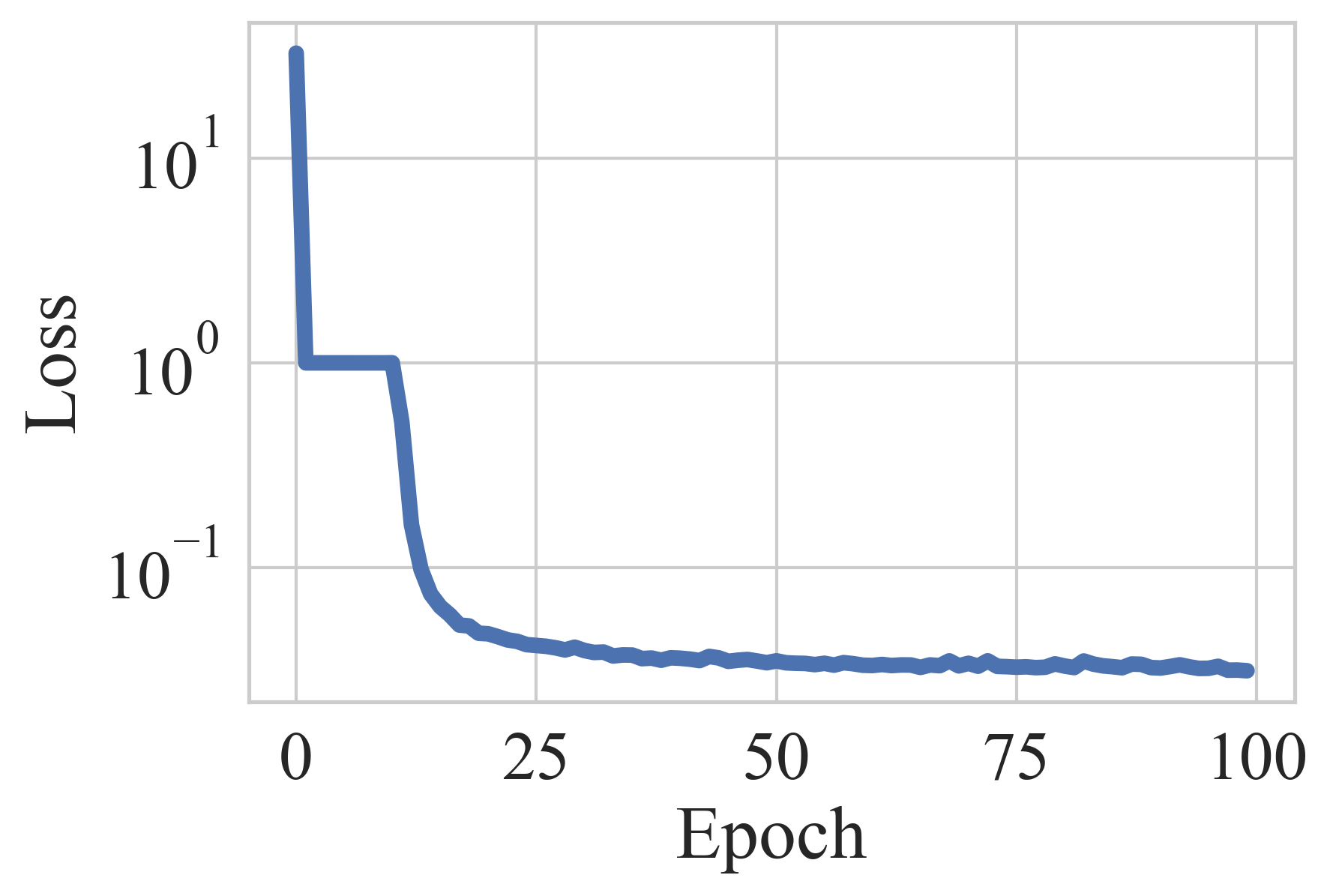} }
\subfigure[8-objective]{ 
\includegraphics[scale=0.27]{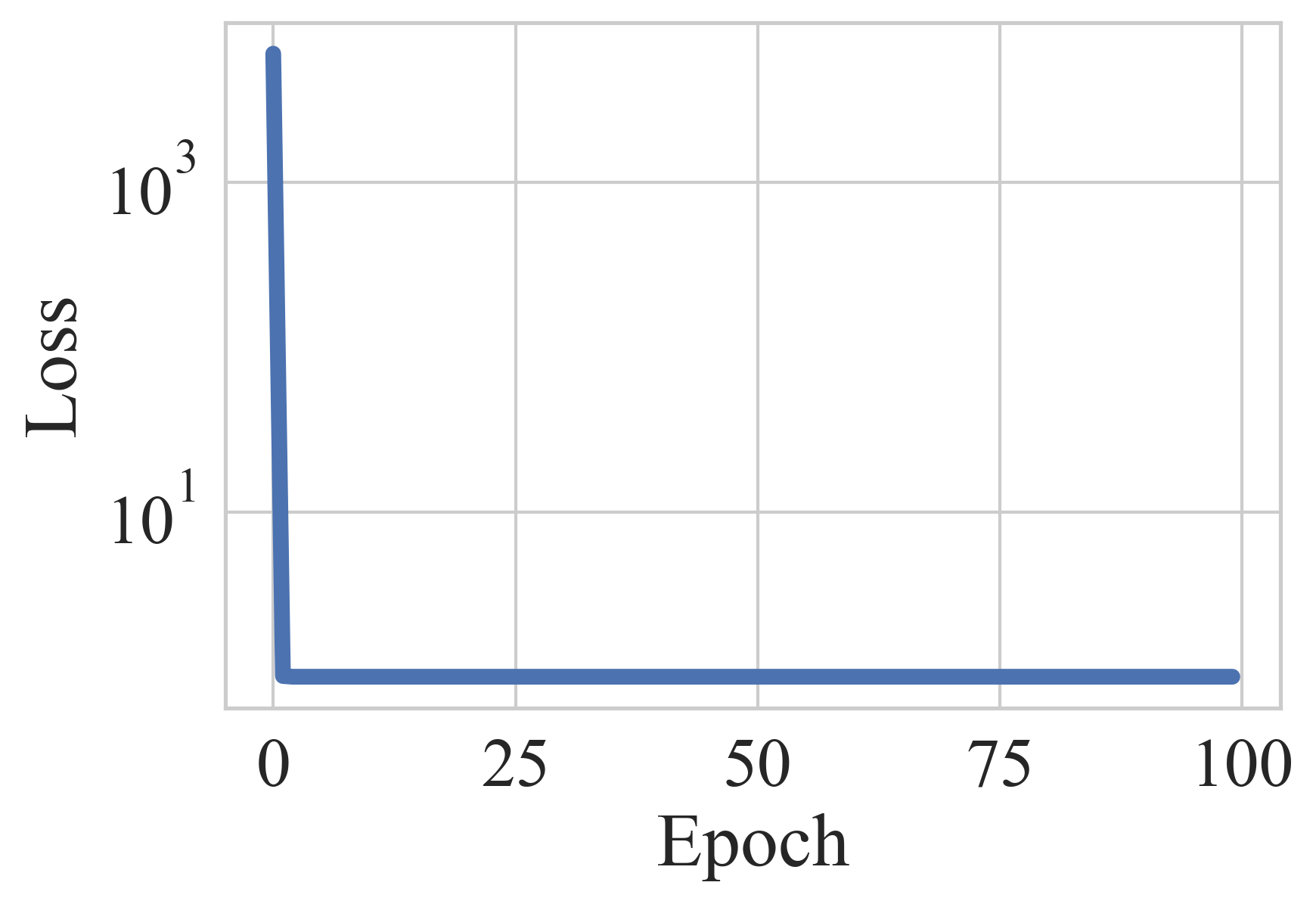} }
\subfigure[10-objective]{
\includegraphics[scale=0.27]{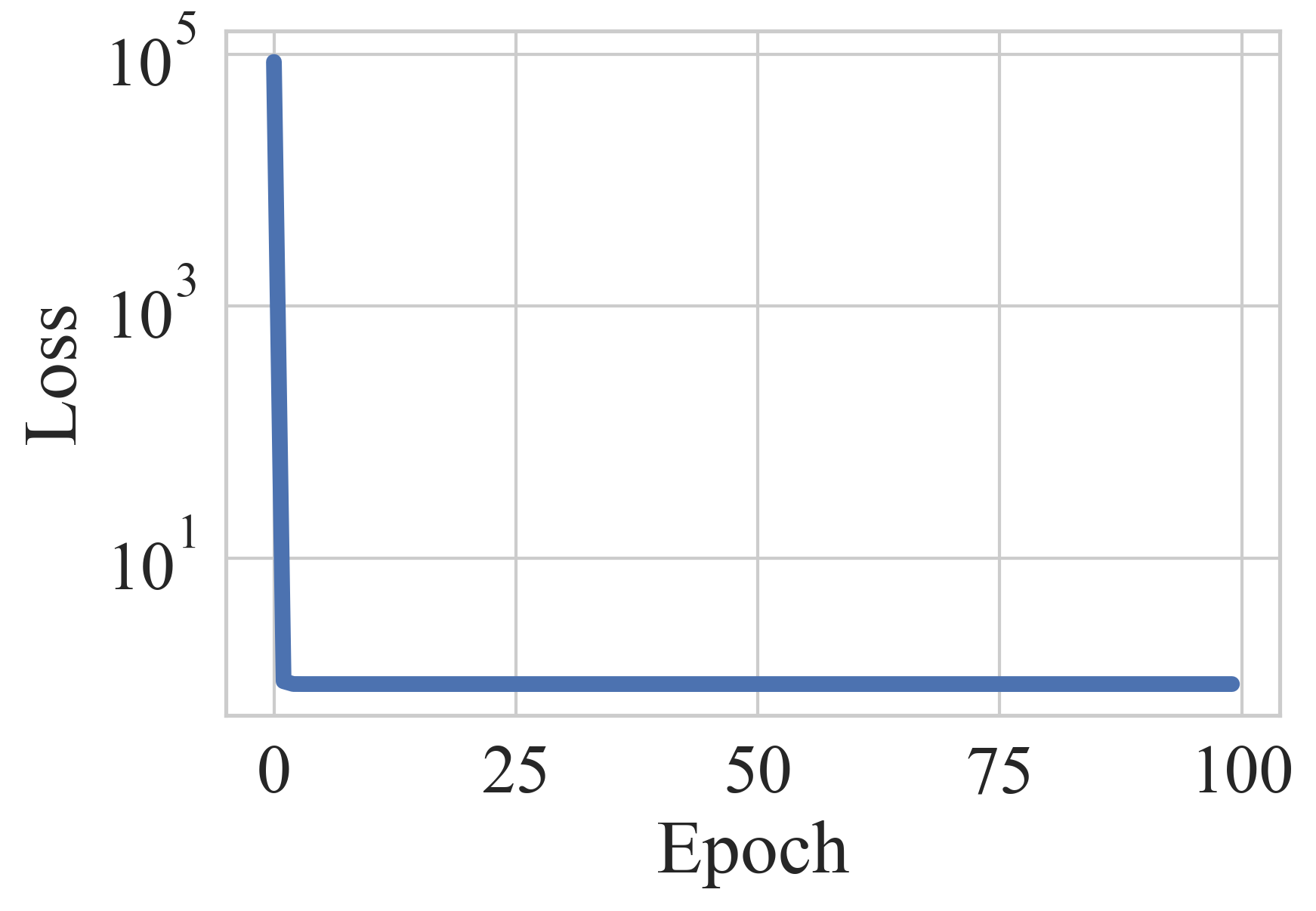} }
\caption{The training curve of HV-Net2 in each case.} 
\label{train1}                                                        
\end{figure}

Fig. \ref{2lossfuns} and Table \ref{table1} show the performance comparison of HV-Net, HV-Net1, and HV-Net2 on the testing solution sets.  We can see that HV-Net1 achieves much worse approximation errors than HV-Net in all cases. HV-Net2 achieves similar approximation errors to HV-Net in 3- and 5-objective cases, but worse approximation errors than HV-Net in 8- and 10-objective cases.  We also show their results on the training solution sets in Table \ref{table2}. We can see that the poor approximation performance on testing data of HV-Net1 and HV-Net2 (in Table \ref{table1}) is due to their poor fitting performance on training data (in Table \ref{table2}). The results in Fig. \ref{2lossfuns} and Tables \ref{table1}-\ref{table2} show the ineffectiveness of using $\mathcal{L}_1$ and $\mathcal{L}_2$ for training HV-Net. The reasons for the ineffectiveness of $\mathcal{L}_1$ and $\mathcal{L}_2$ are explained as follows. 
\begin{enumerate}
\item For $\mathcal{L}_1$, the minimization of $\mathcal{L}_1$ does not mean the minimization of the approximation error $\varepsilon$. If the true and approximated hypervolume values of a solution set are both very small but in different order of magnitude (e.g., true value is $10^{-6}$ and approximated value is $10^{-4}$), we will have a very small $\mathcal{L}_1\approx 10^{-8}$ but a large $\varepsilon\approx 99$. This motivates us to propose $\mathcal{L}$ as the loss function. By calculating the logarithm of each value, we will have a large $\mathcal{L}\approx 21$ which can properly reflect the large approximation error.
\item For $\mathcal{L}_2$, although the minimization of $\mathcal{L}_2$ leads to the minimization of the approximation error $\varepsilon$, it is  difficult to train HV-Net using $\mathcal{L}_2$ in high-dimensional cases. Fig. \ref{train1} shows the training curve of HV-Net2. We can observe that the training loss gets stuck at $10^0$ for 8- and 10-objective cases. This means that the output of HV-Net2 is always very close to zero\footnote{The output of HV-Net2 cannot be zero since the activation function in the output layer is Sigmoid as shown in Fig. \ref{networks}.} for any input solution set, and the parameters of HV-Net2 cannot be further updated to escape from this trap.
\end{enumerate}

We also provide the training curve of HV-Net1 in Section II of the supplementary material. Interested readers can refer there for more information.

\section{Conclusions}
We showed that HV-Net is a promising method for hypervolume approximation. Compared with the point-based and line-based methods, HV-Net achieved better performance in terms of the approximation error and the runtime. We also showed the effectiveness of the proposed loss function over the other two loss functions for training HV-Net. The experimental results are promising and encouraging. We believe that HV-Net can bring new opportunities for the development of the EMO field. 

One disadvantage of HV-Net is that its structure is prespecified before training. Therefore, we can only obtain a single approximation result for each solution set based on a trained HV-Net. This disadvantage elicits our future research: multi-objective HV-Net. That is, we can train a set of HV-Nets with different structures, so that a tradeoff between the approximation error and the runtime can be obtained. The multi-objective neural architecture search technique (e.g., NSGA-Net \cite{lu2019nsga}) can be useful to realize this goal.

\bibliographystyle{IEEEtran}
\bibliography{sample-bibliography}

\end{document}